# MINING OF PRODUCT REVIEWS AT ASPECT LEVEL

Richa Sharma[1], Shweta Nigam[2] and Rekha Jain[3]

[1,2]M.Tech Scholar, Banasthali Vidyapith, Rajasthan, India
[3]Assistant Professor, Banasthali Vidyapith, Rajasthan, India

## ABSTRACT

*Today's world is a world of Internet, almost all work can be done with the help of it, from simple mobile phone recharge to biggest business deals can be done with the help of this technology. People spent their most of the times on surfing on the Web; it becomes a new source of entertainment, education, communication, shopping etc. Users not only use these websites but also give their feedback and suggestions that will be useful for other users. In this way a large amount of reviews of users are collected on the Web that needs to be explored, analyse and organized for better decision making. Opinion Mining or Sentiment Analysis is a Natural Language Processing and Information Extraction task that identifies the user's views or opinions explained in the form of positive, negative or neutral comments and quotes underlying the text. Aspect based opinion mining is one of the level of Opinion mining that determines the aspect of the given reviews and classify the review for each feature. In this paper an aspect based opinion mining system is proposed to classify the reviews as positive, negative and neutral for each feature. Negation is also handled in the proposed system. Experimental results using reviews of products show the effectiveness of the system.*

## KEYWORDS

*Opinion Mining, Sentiment Analysis, Reviews, Aspect, WordNet.*

## 1. INTRODUCTION

There is a rapid growth in a world wide web from the last few years. Diversity of data is available on the web that constitutes the user data. User generated contents include customer reviews, blogs, and discussion forums which expresses customer satisfaction/dissatisfaction on the product and its features explicitly. Large numbers of products are sold and buying on the Web, websites allow their customers to express their opinion on the product that they buy .As the Internet is used by everyone the numbers of reviews that a product receives grow rapidly. This makes it very hard for a potential customer to read them and make a decision on whether to buy the product. Thus, mining this data, identifying the user opinions and classify them is an important task. Sentiment Analysis is a natural language processing task that deals with finding orientation of opinion in a piece of text with respect to a topic [4].

Three main components of Opinion Mining are:

1. Opinion Holder: Person that expresses the opinion is opinion holder.
2. Opinion Object: Object on which opinion is given.
3. Opinion Orientation: Determine whether the opinion about an object is positive, negative or neutral.

For example "The sound quality of this computer is good". In this review, Opinion Holder is the user who has written this review. Opinion object here is the "sound quality" of the computer and the opinion word is "good" which is positively orientated. Semantic orientation is a task of determining whether a sentence has either positive, negative orientation or neutral orientation [6][14].

Opinion mining is performed at three levels [3]:

- **Document level**: At this level the whole document is classified as positive, negative or neutral.
- **Sentence level:** At this level the whole sentence is classified as positive, negative or neutral.
- **Aspect level:** At this level the whole document/sentence is classified as positive, negative or neutral for each feature present in the document/sentence.

Document level and sentence level only classify the whole document or sentence, it does not identify the aspect present in the document/sentence i.e. if the polarity of the document is positive/negative it doesn't mean that document possess positive/negative opinion for each aspect. To determine the opinion on every aspect, opinion mining at aspect level is performed. Two types aspects are found in user reviews explicit and implicit [2].

- **Explicit aspect**s are those that are easily identified in the reviews, explicit aspects are noun and noun phrase. For example, "*The voice quality of this phone is great,* here *voice quality* is an explicit aspect and it can be directly seen in the sentence.
- **Implicit aspects** are those that are not easily identified in the reviews, explicit aspects are not noun and a noun phrase. For example, "*This phone is not fit in my pockets",* here *not fit in my pockets* is an implicit aspect it indicate the aspect size that cannot be directly identified in the sentence.

In this paper an Aspect based Opinion Mining system named as "Aspect based Sentiment Orientation System" is proposed which extracts the feature and opinions from sentences and determines whether the given sentences are positive, negative or neutral for each feature. Negation is also handled by the system. To determine the semantic orientation of the sentences a dictionary based technique of the unsupervised approach is adopted. To determine the opinion words and their synonyms and antonyms WordNet is used as a dictionary. The rest of the paper is organized as follows: Section 2 discusses related work. Section 3 describes the proposed approach. Section 4 shows the experimental results of the system. Section 5 concludes the paper.

## 2. EXISTING RESEARCH WORK

Some of the existing researches in the aspect based opinion mining are mentioned below: By using a distance based approach Hu and Liu [11] extract opinion words and phrases after extracting aspects. To calculate the polarity of each extracted opinion word WordNet was used. 10 "syntactic dependency rule templates" were used by A.-M. Popescu et al. [1] over a dependency tree to relate identified product features to the opinion words. Relaxation-labeling technique to determine the semantic orientation of potential opinion words in the given review. The negative words were also handled by them.

Sentiment lexicon was constructed by N. Godbole et al. [12] by using a Word Net, and sentiments were associated with each entity and it was assumed that a sentiment word found in the same sentence as an entity. A propagation based method was proposed by Qiu et al. [6] to extract opinion and aspects by using bootstrapping approach. This method is based on the idea that the natural relation exists between opinion words and aspects because opinion words are used to describe aspects.

International Journal in Foundations of Computer Science & Technology (IJFCST), Vol.4, No.3, May 2014By using association rule mining Hu and Liu [11] extract aspects from customer reviews. To find the frequent features pruning strategies and association rule mining was employed A.-M. Popescu [1] removed the noun phrases that are not features by calculating a point wise mutual information (PMI) score for each noun phrase. J. Yi et al. [9] built a system for opinion extraction based on unsupervised technique .They developed and tested two feature term selection algorithms based on a mixture language model and likelihood ratio.

Kessler and Nicolov [10] used a supervised learning method i.e SVM (Support Vector Machine) to find the aspect and opinion and identify which aspect is related to which opinion expression in a review. Classifier compared by the algorithm is given by Bloom et al. and it showed better performance based on F-measure. Machine learning approach is used by W. Jin et al. [18] which is built under the frame of lexicalized Hidden Markov Model (LHMMs). They called this approach "Opinion Miner".

Conditional Random Fields (CRF)-based approach used by N. Jakob et al. [13] for opinion target extraction. Several features were used as an input such as, short dependency path, POS tags, word distance and opinion sentence. To evaluate this method datasets from different sources employed and it has been showed that this method can be effectively used in single-domain and cross domain setting. V. Stoyanov et al. [17] treated this task as a topic co reference resolution problem and used supervised approach to perform this task. Opinions that share the same target are clustered together.

## 3. PROPOSED SYSTEM

Proposed system is based on unsupervised technique [4]. Dictionary based approach of the unsupervised technique is used to determine the orientation of sentences. WordNet [7] is used as a dictionary to determine the opinion words and their synonyms and antonyms. The proposed work is closely related to the Minqing Hu and Bing Liu work on Mining and Summarizing Customer Reviews [11]. Figure1. Gives the overview of the proposed system 'Aspect based Sentiment Orientation System' [15][16]. The reviews of mobile phones are considered as an input to the system all the reviews are collected from the Amazon website (www.amazon.in) which is used as an input to the system. All the features of the product on which reviews are given would be identified and the orientation of the sentence for each feature would be determined. The polarity of the given sentence is determined on the basis of the majority of opinion words. In the end the system will generate the feature wise summary of positive, negative and neutral sentences which will be easier for users to read, analyze and help them in taking the decision whether the product is to be purchased or not.

The system performs this task in several steps as follows:-

### 3.1 Data Collection

To determine the polarity of the sentences, based on aspects, large numbers of reviews are collected from the Web. There are lots of websites on the Internet where the large numbers of customer reviews are available. Amazon website (www.amazon.com) is used to collect the reviews.

### 3.2 POS Tagging

After collecting the reviews, they are sent to the POS tagging module where POS tagger tag all the words of the sentences to their appropriate part of speech tag [5][8]. POS tagging is an important phase of opinion mining, it is necessary to determine the features

89

and opinion words from the reviews.POS tagging can be done manually or with the help of POS tagger. Manual POS tagging of the reviews take lots of time. Here, POS tagger is used to tag all the words of reviews.

### 3.3 Feature Extraction

All the features are extracted from the reviews and stored in a database then its corresponding opinion words are extracted from these reviews.

### 3.4 Extracting Opinion Words & Seed List Preparation

Initially some of the common opinion words along with their polarity are stored in the seed list. All the opinion words are extracted from the tagged output. The extracted opinion words matched with the words stored in seed list .If the word is not found in the seed list then the synonyms are determined with the help of WordNet. Each synonym is matched with words in the seed list, if any synonym matched then extracted opinion word is stored with the same polarity in the seed list. If none of the synonym is matched then the antonym is determined form the WordNet and the same process is repeated, if any antonym matched then extract opinion word is stored with the opposite polarity in the seed list. In this way the seed list keep on increasing. It grows every time whenever the synonyms or antonyms words are found in WordNet matches with seed list.

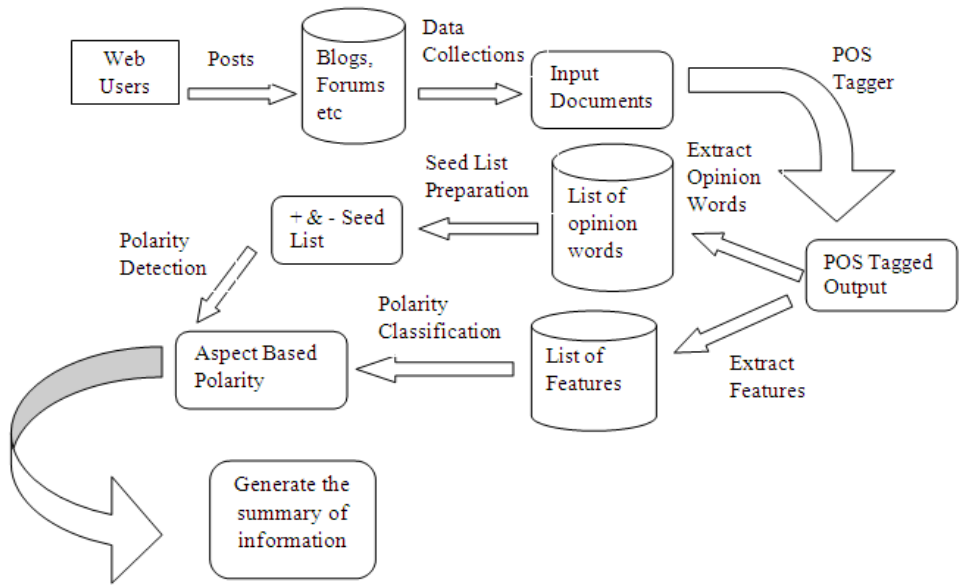

Figure 1 Aspect based sentiment orientation system

### 3.5 Polarity Detection & Classification

With the help of seed list, the polarity of the sentences is determined for each feature. Polarity is determined on the basis of majority of opinion words, if the number of positive words is more, then the polarity of the sentence is positive otherwise the polarity is negative and if the number of positive and negative words is equal then the sentence shows the neutral polarity. Negation is also handled in the system, if the opinion word is preceded by negation "not" then polarity of that sentence is reversed. For example, the sentence *"The touch screen of this mobile phone is not good"* shows the negative polarity because the opinion word '**good**' is preceded by '**not**'. In the end, a features wise summary is generated. Figure 2 shows how the feature wise summary is generated by the system.

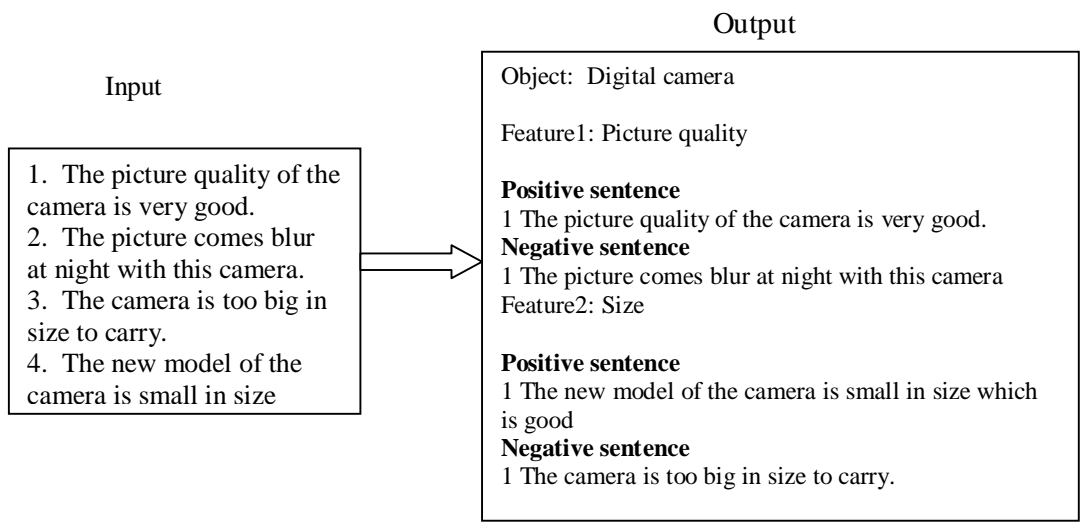

Figure 2 Feature wise summary generated by the system

System does not handle the repetition of the same information. For example, If the same review is repeated two times then the system classify the review two times as positive, negative or neutral. But, if the words having same meaning or same words are repeated in a sentence then they are counted separately. For example. The sentence, *"The camera of this phone is good and the night mode effect is excellent but the flash is bad".* In this sentence the words *'good'* and *'excellent'* both have the same meaning, 'good' is the synonyms of 'excellent' in the wordnet hierarchy. *Good* and *Excellent* both shows the positive polarity ,they are counted as two positive words in a sentence and ***Bad*** shows the negative polarity which is counted as one negative word in the sentence. The overall polarity of the sentence is positive for the feature 'Camera'.

## 4. EXPERIMENTAL RESULTS

Customer reviews of mobile phones are used for the experimental purpose and collected from the Amazon.com (www.amazon.in). Collected reviews of the mobile phones are applied to the system. The result shows the orientation of each sentences i.e. whether a sentence is positive, negative or neutral for each feature that reviews contain. The final results are shown in graphical charts. The results of current system are compared with human decision that helps to evaluate the current system. All the reviews are read manually first and their corresponding opinion is determined. The results are then compared with the results of "Aspect based Sentiment orientation system" Three evaluation measures are used on the basis of which systems are compared, these are:-

- Precision
- Recall
- Accuracy

On the basis of these evaluation measures, results show that 'Aspect based Sentiment Orientation System' is performed well in phone domain. The experiments have been performed by using 50 sentences of phone reviews.

Figure 3 present the results of **'**Feature based system performance' which shows the precision, recall and accuracy of each feature in graphical form.

Figure 4 presents the results of 'Feature wise error rate and correct rate'. i.e. how many sentences system has classified correctly and how many sentences system has classified wrong on the basis of each feature.

The overall performance of the system has shown in the form of pie chart in Figure 5.
Figure 6 represents the overall system's error and correct rate it shows how many sentences are categorized correctly with correct polarity.

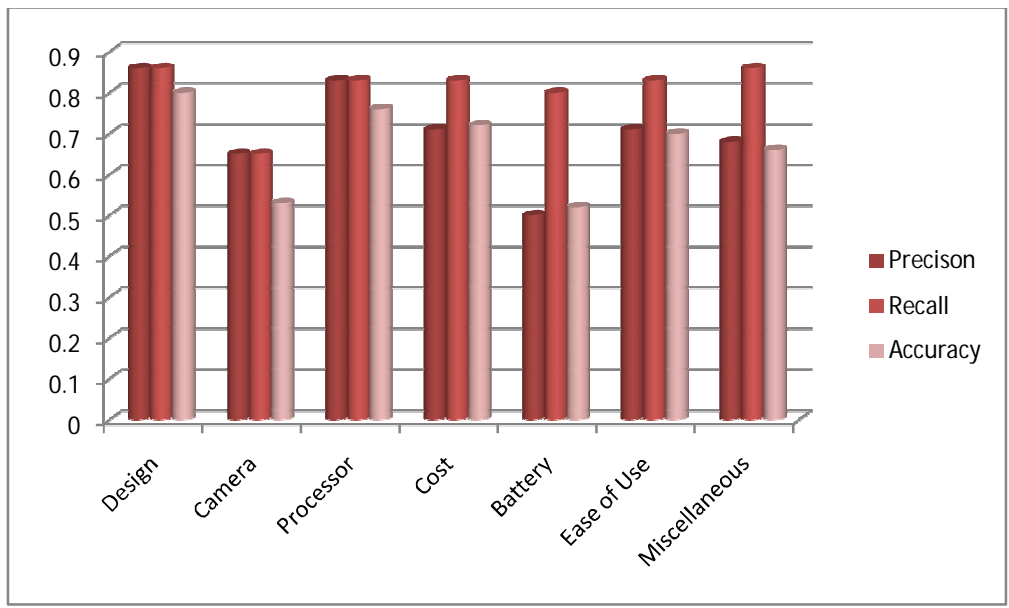

Figure 3. Feature based system performance graph

International Journal in Foundations of Computer Science & Technology (IJFCST), Vol.4, No.3, May 2014

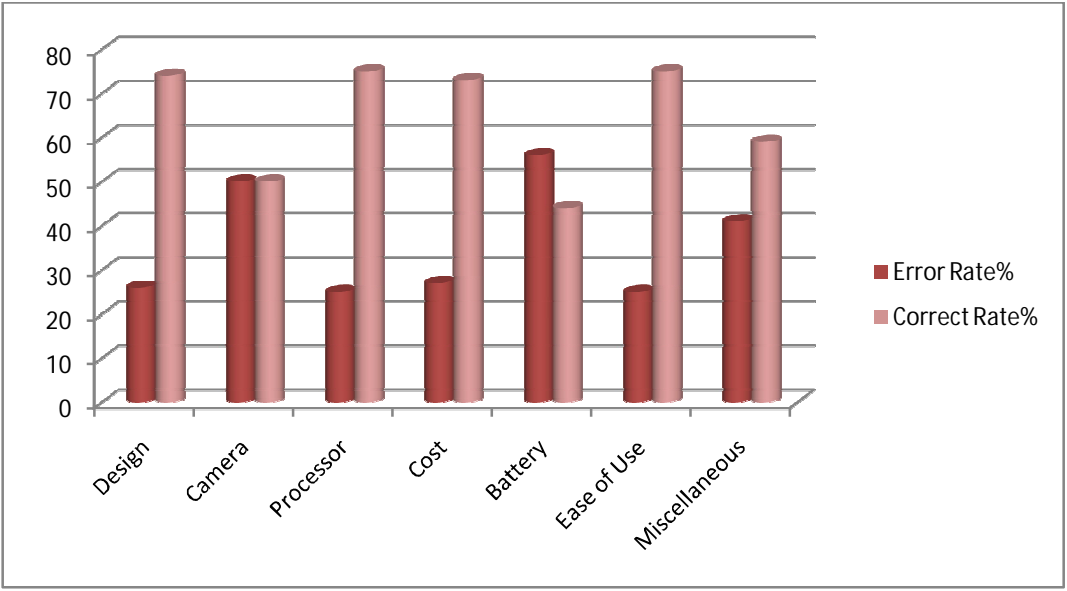

Figure 4. Feature wise error and correct rate graph

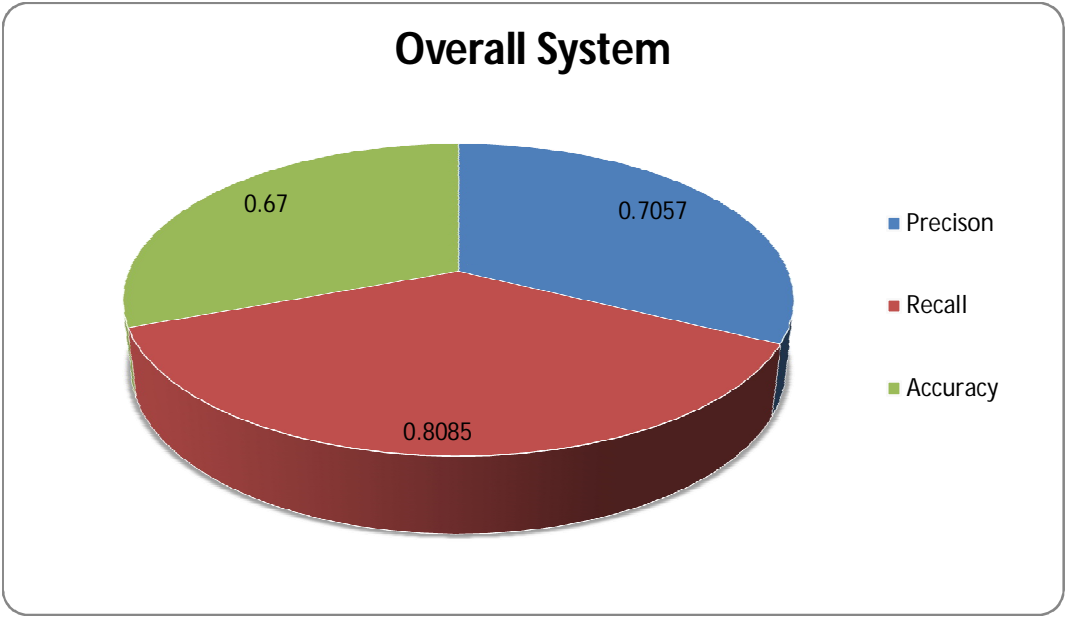

Figure 5 Overall System performance graph

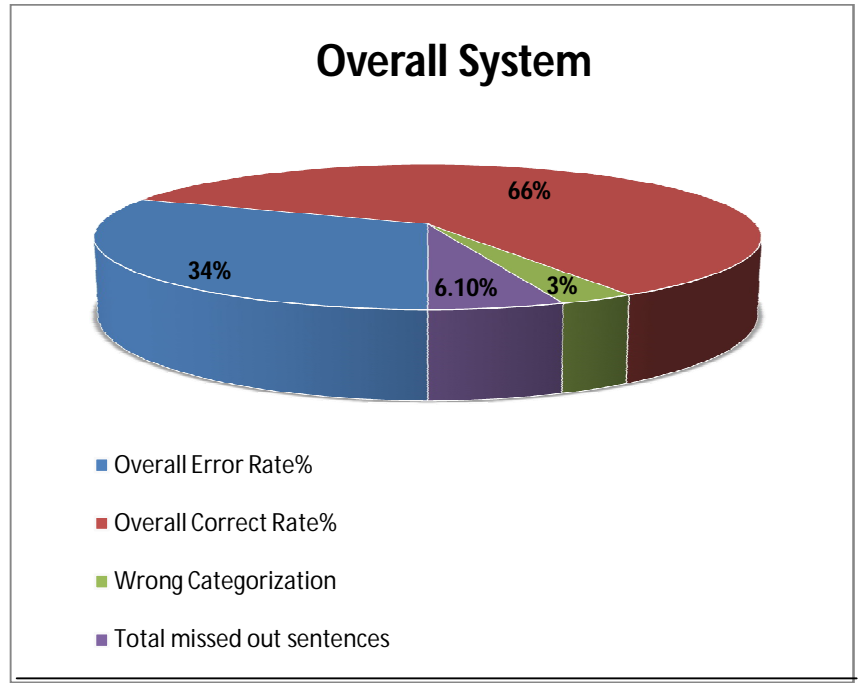

Figure 6. Overall System error and correct rate graph

The above results show that the 'Aspect based Sentiment Orientation System' performs well with respect to the phone domain and shows the accuracy of 67% which proves the system is efficient.

## 5. CONCLUSION

The objective of this paper is to determine the polarity of the customer reviews of mobile phones at aspect level. System performs the aspect based opinion mining on the given reviews and the feature wise summarized results generated by the system will be helpful for the user in taking the decision .Experimental results indicate that the 'Aspect based Sentiment Orientation System' perform well and has achieved the accuracy of 67%.Aspect based opinion mining is necessary because nowadays everyone is busy and they don't have a time to read all the positive or negative reviews  if someone just wants to know about some feature of the product. Aspect based opinion mining has proved to be helpful in these situations as compared to simple opinion mining. In future work, efforts would be done to make some enhancements in this technique in such a way that it can identify the repeated reviews and classify those reviews only once. It would deal with the sentences contain relative clauses like not only-but also and the sentences contain clauses neither-nor, either-or etc. We also plan to perform opinion mining in Hindi language.

International Journal in Foundations of Computer Science & Technology (IJFCST), Vol.4, No.3, May 2014

[4] Bo Pang, Lillian Lee,(2008)"Opinion mining and sentiment analysis". Foundations and Trends in Information Retrieval, Vol. 2(1-2):pp. 1–135.
[5] Christopher D. Manning ,"Part-of-Speech Tagging from 97% to 100%: Is It Time for Some Linguistics?",Published in CICLing'11 Proceedings of the 12th international conference on Computational linguistics and intelligent text processing - Volume Part I.
[6] G. Qiu, B. Liu, J. Bu, and C. Chen,(2011) "Opinion word expansion and target extraction through double propagation," Comput. Linguist., vol. 37, pp. 9-27.
[7] George A. Miller, Richard Beckwith, Christiane Fellbaum, Derek Gross, and Katherine Miller (1990)," Introduction to WordNet: An On-line Lexical Database (Revised August 1993) International Journal of Lexicography" 3(4):235-244.
[8] Kevin Gimpel, Nathan Schneider, Brendan O 'Connor, Dipanjan Das, Daniel Mills,Jacob Eisenstein, Michael Heilman, Dani Yogatama, Jeffrey Flanigan, and Noah A. Smith, " Part-of-Speech Tagging for Twitter: Annotation, Features, and Experiments", Published in Proceeding HLT '11 Proceedings of the 49th Annual Meeting of the Association for Computational Linguistics: Human Language Technologies: short papers - Volume 2  ISBN: 978-1-932432-88-6
[9] J. Yi, T. Nasukawa, R. Bunescu, and W. Niblack,(2003), "Sentiment Analyzer: Extracting Sentiments about a Given Topic using Natural Language Processing Techniques," presented at the Proceedings of the Third IEEE International Conference on Data Mining.
[10] J. S. Kessler and N. Nicolov,(2009), "Targeting sentiment expressions through supervised ranking of linguistic configurations," in Proceedings of the Third International AAAI Conference on Weblogs and Social Media,,San Jose, California, USA, pp. 90-97.
[11] M. Hu and B. Liu, "Mining and summarizing customer reviews," presented at the Proceedings of the tenth ACM.
[12] N. Godbole, M. Srinivasaiah, and S. Skiena, (2007),"Large-scale sentiment analysis for news and blogs," in International Conference on Weblogs and Social Media (ICWSM), pp. 219-222.
[13] N. Jakob and I. Gurevych,(2010), "Extracting opinion targets in a single- and cross-domain setting with conditional random fields," presented at the Proceedings of the 2010 Conference on Empirical Methods in Natural Language Processing, Cambridge, Massachusetts.
[14] P. Turney, "Thumbs up or thumbs down? Semantic orientation applied to unsupervised classification of reviews," Proceedings of the Association for Computational Linguistics.
[15] Richa Sharma, Shweta Nigam, Rekha Jain,(2014), Polarity detection at sentence level. International journal of computer applications(0975 – 8887), Volume 86 – No 11.
[16] Richa Sharma, Shweta Nigam, Rekha Jain,(2014), Determination of Polarity of sentences using Sentiment Orientation System. International journal of Advances in Computer Science and Technology (IJACST) WARSE, Volume 3, No.3, March 2014.
[17] V. Stoyanov and C. Cardie,(2008) "Topic identification for finegrained opinion analysis" ,presented at the Proceedings of the 22nd International Conference on Computational Linguistics Volume 1, Manchester, United Kingdom.
[18] W. Jin, H. H. Ho, and R. K. Srihari,(2009), "OpinionMiner: a novel machine learning system for web opinion mining and extraction," presented at the Proceedings of the 15thACM SIGKDD international conference on Knowledge discovery and data mining, Paris, France.